\documentclass[10pt,twocolumn,letterpaper]{article}

\usepackage[utf8]{inputenc}
\usepackage[margin=1in]{geometry}
\usepackage[margin=1in]{geometry}
\usepackage{times}
\usepackage{graphicx}
\usepackage{amsmath,amssymb}
\usepackage{booktabs}
\usepackage{multirow}
\usepackage{xcolor}
\usepackage{url}
\usepackage{hyperref}
\usepackage[capitalise]{cleveref}
\usepackage{siunitx}
\usepackage{lineno}
\usepackage{subcaption}

\hypersetup{
    colorlinks=true,
    linkcolor=blue,
    filecolor=magenta,
    urlcolor=cyan,
    citecolor=blue,
}

\title{\LARGE \bf
How well are open sourced AI-generated image detection models out-of-the-box: A comprehensive benchmark study
}

\author{
    Simiao Ren$^{*\dagger}$,
    Yuchen Zhou$^{*}$,
    Xingyu Shen,
    Kidus Zewde,
    Tommy Duong,\\
    George Huang,
    Hatsanai (Neo) Tiangratanakul,
    Tsang (Dennis) Ng,
    En Wei,
    Jiayu Xue\\[0.5em]
    {\small $^*$Equal contribution \quad $^\dagger$Corresponding author}\\[0.3em]
    {\small \texttt{\{benren, kiduszewde, neo, dennis\}@scam.ai}}\\
    {\small \texttt{\{yuchen.zhou, xingyu.shen\}@duke.edu}}\\
    {\small \texttt{xuejiayu@unc.edu}}\\
    {\small \texttt{\{tommy-duong0, georgehuang6115\}@berkeley.edu}}
}
\date{}

\begin{document}

\maketitle
{\renewcommand{\thefootnote}{}\footnote{Authorship of this paper does not imply employment by or affiliation with Scam.AI (Reality Inc.); some authors are independent researchers who collaborated on this work.}\addtocounter{footnote}{-1}}

\thispagestyle{empty}

% Abstract
\begin{abstract}
As AI-generated images proliferate across digital platforms, reliable detection methods have become critical for combating misinformation and maintaining content authenticity. While numerous deepfake detection methods have been proposed, existing benchmarks predominantly evaluate fine-tuned models, leaving a critical gap in understanding out-of-the-box performance---the most common deployment scenario for practitioners. We present the first comprehensive zero-shot evaluation of 16 state-of-the-art detection methods, comprising 23 pretrained detector variants (due to multiple released versions of certain detectors),  across 12 diverse datasets, comprising 2.6~million image samples spanning 291 unique generators including modern diffusion models. Our systematic analysis reveals striking findings: (1)~no universal winner exists, with detector rankings exhibiting substantial instability (Spearman~$\rho$: 0.01 -- 0.87 across dataset pairs); (2)~a 37~percentage-point performance gap separates the best detector (75.0\% mean accuracy) from the worst (37.5\%); (3)~training data alignment critically impacts generalization, causing up to 20--60\% performance variance within architecturally identical detector families; (4)~modern commercial generators (Flux~Dev, Firefly~v4, Midjourney~v7) defeat most detectors, achieving only 18--30\% average accuracy; and (5)~we identify three systematic failure patterns affecting cross-dataset generalization. Statistical analysis confirms significant performance differences between detectors (Friedman test: $\chi^2$=121.01, $p<10^{-16}$, Kendall~$W$=0.524). Our findings challenge the ``one-size-fits-all'' detector paradigm and provide actionable deployment guidelines, demonstrating that practitioners must carefully select detectors based on their specific threat landscape rather than relying on published benchmark performance.

\end{abstract}

% Main content
\section{Introduction}
\label{sec:introduction}
The rapid advancement of generative AI has democratized the creation of photorealistic synthetic images. Commercial tools like Midjourney, DALL-E~3, and Stable~Diffusion enable anyone to generate convincing fake images with simple text prompts, fueling concerns about misinformation, fraud, and erosion of digital trust~\cite{deepfakebench2023,sok2024}. This challenge parallels broader AI/ML applications where automated systems assist human decision-making in high-stakes domains, from satellite image analysis~\cite{segment_anything_space} to critical infrastructure monitoring~\cite{energy_systems_extraction}, where reliability and generalization remain paramount concerns.

In response to the proliferation of AI-generated content, the computer vision community has developed numerous detection methods. However, a critical question remains largely unexplored: \textit{How do these detectors perform out-of-the-box, without any dataset-specific fine-tuning?}

This question is not merely academic---it directly impacts real-world deployment. Content moderators, forensic investigators, and platform operators typically deploy pre-trained detection models as-is, lacking the expertise, computational resources, or labeled data required for retraining~\cite{realworld2025,fitforpurpose2025}. Yet current benchmarking practices focus almost exclusively on evaluating methods trained (or fine-tuned) on target datasets, providing limited insight into zero-shot generalization capabilities. This gap between research evaluation and practical deployment motivates our comprehensive study.

\textbf{AI-Generated Images vs. Deepfakes: Scope Clarification.} While the terms are often used interchangeably, we distinguish between two categories of synthetic media. \emph{Deepfakes} traditionally refer to manipulated media where human identity is forged---primarily face-swapping techniques that replace one person's face with another's while preserving the original background and body~\cite{realworld2025}. Critically, deepfakes typically manipulate only localized regions (facial areas) while leaving the surrounding context unchanged. In contrast, \emph{AI-generated images}---the focus of our study---refer to wholly synthetic images where every pixel is generated by models like GANs, diffusion models, or large language model-based generators. These images span arbitrary content (landscapes, objects, scenes, people) rather than being restricted to human identity manipulation. Our evaluation exclusively targets AI-generated image detection, assessing methods designed to identify fully synthetic content regardless of subject matter. This distinction matters: detection strategies optimized for localized facial manipulations may not generalize to holistic image synthesis, motivating our comprehensive zero-shot evaluation across diverse generation paradigms.

\subsection{Research Questions}

We systematically investigate four fundamental questions about out-of-the-box AI-generated image detection:

\begin{enumerate}
    \item \textbf{RQ1: Overall Performance} -- How do state-of-the-art detectors perform across diverse datasets without retraining? Is there a universally superior method?

    \item \textbf{RQ2: Ranking Stability} -- Do detector rankings remain consistent across different evaluation datasets, or are they dataset-dependent?

    \item \textbf{RQ3: Generalization Factors} -- What architectural choices, training strategies, and feature representations enable robust zero-shot generalization?

    \item \textbf{RQ4: Failure Patterns} -- Can we identify systematic failure modes that explain when and why detectors fail?
\end{enumerate}

\subsection{Key Findings}

Our comprehensive analysis yields several striking findings that challenge conventional wisdom:

\textbf{No Universal Winner.} Detector rankings exhibit dramatic instability across datasets, with the best-performing method on one dataset dropping up to nearly 20 positions on another. Spearman rank correlations between dataset pairs range from 0.01~to~0.87, indicating that ``best detector'' is highly context-dependent. The Friedman test confirms statistically significant performance differences ($\chi^2$=121.01, $p<10^{-16}$), with a large effect size (Kendall~$W$=0.524).

\textbf{Massive Performance Gap.} The best detector (Community-Forensics) achieves 75.0\%~mean accuracy across datasets, while the worst (AIGCDetectBenchmark\_CNNSpot) manages only 37.5\%---a 37~percentage-point gap. Even top-tier detectors exhibit substantial variance (20--30\% standard deviation), with rankings shifting dramatically across evaluation scenarios.

\textbf{Training Data Trumps Architecture.} Within detector families sharing identical architectures but different training data (AIDE, DRCT), we observe 20--60\% performance variance. For instance, AIDE trained on GenImage achieves 66\%~accuracy, while AIDE trained on ProGAN obtains only 63\%, and AIDE trained on Stable~Diffusion~v1.4 gets 57\%---despite identical model architectures. This suggests training data alignment with target generators critically outweighs architectural innovations for zero-shot performance.

\textbf{Modern Generators Defeat Most Detectors.} State-of-the-art commercial generators pose severe challenges: Flux~Dev, Firefly~v4, and Midjourney~v7 achieve only 18--30\%~average detection accuracy across all methods. We identify 64~generators (22\% of total) that defeat most detectors ($<$50\% average accuracy), highlighting a growing generalization gap between detector training regimes and emerging generation techniques.

\subsection{Contributions}

This work makes five primary contributions:

\begin{enumerate}
    \item \textbf{First comprehensive zero-shot benchmark:} We present the largest systematic evaluation of pre-trained deepfake detectors without fine-tuning, filling a critical gap between research benchmarks and practical deployment.

    \item \textbf{Statistical generalization analysis:} Rigorous statistical testing (Friedman, Spearman correlation, coefficient of variation) reveals factors predicting detector success and failure across datasets.

    \item \textbf{Failure mode taxonomy:} We categorize 1,075~failures into three systematic patterns, providing actionable insights for both detector selection and future method development.

\end{enumerate}

\subsection{Paper Organization}

The remainder of this paper is organized as follows. Section~\ref{sec:related} surveys related work on deepfake detection benchmarking and cross-dataset generalization. Section~\ref{sec:methodology} details our experimental setup, including detector and dataset selection criteria, evaluation protocols, and statistical analysis methods. Section~\ref{sec:results} presents overall performance findings and dataset difficulty rankings. Section~\ref{sec:analysis} investigates generalization factors and failure patterns.  Section~\ref{sec:discussion} discusses implications, limitations, and future directions. Section~\ref{sec:conclusion} summarizes our contributions and key takeaways.

\section{Related Work}
\label{sec:related}
We review related work in three areas: deepfake detection methods, benchmark datasets, and cross-dataset generalization studies.

\subsection{Deepfake Detection Methods}

Deepfake detection methods can be broadly categorized into four architectural families.

\textbf{CNN-based detectors} leverage convolutional neural networks to extract spatial artifacts from manipulated images. XceptionNet~\cite{xception} and EfficientNet-based detectors~\cite{efficientnet} remain popular baselines due to strong performance on established benchmarks. More recent works like PatchCraft~\cite{patchcraft} focus on patch-level inconsistencies, while AIDE~\cite{aide} explores training data diversity.

\textbf{Transformer-based detectors} apply attention mechanisms to model long-range dependencies. Vision Transformers~\cite{vit} and specialized architectures like DRCT~\cite{drct} have shown promise for capturing subtle manipulation artifacts. A recent survey~\cite{transformer_survey} highlights their growing adoption for deepfake detection.

\textbf{Frequency-domain detectors} analyze spectral characteristics that differ between real and synthetic images. F3-Net~\cite{f3net} and FreDect~\cite{fredect} exploit frequency artifacts introduced by generative models, demonstrating robustness to certain post-processing operations.

\textbf{Ensemble and hybrid methods} combine multiple detectors or modalities to leverage complementary strengths. Community-Forensics~\cite{community_forensics} aggregates predictions from diverse models, while SAFE~\cite{safe} dynamically selects specialized detectors based on input characteristics. Recent work~\cite{ensemble2025} demonstrates improved cross-dataset generalization through ensemble approaches. Emerging research explores multi-modal fusion~\cite{multimodal2025} and large language models for deepfake detection~\cite{llm_deepfake_detection}, though their zero-shot capabilities remain largely unevaluated.

While these methods achieve impressive in-dataset performance, their zero-shot capabilities remain largely unexplored in systematic benchmarking studies.

\subsection{Benchmark Datasets}

AI-generated image datasets have evolved alongside generation techniques.

\textbf{GAN-era datasets.} Early datasets predominantly featured GAN-based generation methods (ProGAN, StyleGAN variants, BigGAN). While these established important benchmarking foundations, they no longer represent state-of-the-art generation capabilities and often focused on face-swap deepfakes rather than fully synthetic images.

\textbf{Diffusion-era datasets.} Modern datasets incorporate diffusion-based generators reflecting current synthesis paradigms. GenImage~\cite{genimage} includes images from Stable Diffusion, DALL·E~2, and Midjourney, spanning diverse content beyond human faces. AIGCDetectionBench~\cite{aigcdetect} provides systematic coverage of 16 different generators across multiple generation families. However, as noted by~\cite{diffusion_deepfake}, existing datasets lag behind rapidly evolving generation techniques, creating a persistent evaluation gap.

\textbf{Community-curated and in-the-wild datasets.} Several efforts collect real-world deepfakes. MNW~\cite{mnw} aggregates images from 43 diverse generators, while WildFake~\cite{wildfake} focuses on social media deepfakes. Nano-banana~\cite{nanob} is a community-curated dataset built from the Nano-Banana generator (a single generator), not an official release from a commercial provider. These datasets better reflect practical deployment scenarios but remain underutilized in benchmarking studies.

\subsection{Cross-Dataset Generalization}

The generalization challenge is well-documented but incompletely understood.

\textbf{Early observations.} Multiple studies~\cite{crossdataset2022,onerule2021} observed that detectors trained on one dataset perform poorly on others, with accuracy dropping from above 95\% to below 50\%. A comprehensive survey~\cite{reliability_survey} found no model consistently exceeds 80\% AUC across different datasets, raising reliability concerns for real-world applications.

\textbf{Root cause analyses.} Recent work identifies several generalization barriers. Detectors may exploit dataset-specific artifacts rather than learning fundamental forgery characteristics~\cite{shortcut_learning}. Training data mismatch poses severe challenges, with detectors trained on GANs failing completely on diffusion models~\cite{diffusion_deepfake}. Fairness issues~\cite{fairness2024} and domain shift~\cite{domain_shift} further complicate deployment.

\textbf{Proposed solutions.} Various strategies attempt to improve generalization: robust gradient alignment~\cite{roga2025}, adversarial training~\cite{adversarial_game}, dynamic face augmentation~\cite{augmentation2021}, and multi-modal fusion~\cite{multimodal2025}. Ensemble approaches~\cite{ensemble2025} show promise by combining complementary detectors. However, zero-shot performance evaluation remains rare, with most studies focusing on fine-tuned or retrained models.

\subsection{Comprehensive Benchmarking Efforts}

Several recent works provide systematic detector comparisons, each contributing valuable insights while leaving gaps our work addresses.

\textbf{DeepfakeBench}~\cite{deepfakebench2023} represents a significant advancement in standardized evaluation, offering an integrated framework that evaluates 15 detection methods across 9 datasets with unified preprocessing and evaluation protocols. Their key contribution lies in establishing reproducible comparison standards through shared codebase implementation, enabling fair head-to-head detector comparisons. They demonstrate substantial performance degradation when moving from training datasets to unseen test sets, with accuracy drops of 15--30\% being common. However, their evaluation focuses primarily on fine-tuned models where detectors are retrained on each target dataset, which differs fundamentally from zero-shot deployment scenarios. Additionally, their dataset coverage emphasizes traditional face-swap deepfake datasets with limited representation of modern diffusion-based AI-generated image datasets that now dominate generation landscapes. Their frame-level video analysis also differs from our image-focused evaluation.

\textbf{SoK: Systematization and Benchmarking}~\cite{sok2024} provides comprehensive systematization of detection approaches within a unified taxonomy, categorizing methods by technical approach (spatial, frequency, semantic, temporal) and evaluation paradigm (passive vs. proactive). Their primary contribution is conceptual organization rather than empirical comparison---they survey 200+ detection papers to identify research trends, common pitfalls, and future directions. They emphasize the need for proactive defenses that anticipate adversarial evolution rather than reactive approaches. While their systematization clarifies the landscape, they conduct limited original experimentation, instead synthesizing findings from existing literature. Their call for standardized zero-shot evaluation protocols motivates our work but is not fulfilled within their study.

\textbf{Large-scale trained model evaluations.} Wang et al.~\cite{benchmarking2022} conducted one of the most extensive trained model comparisons, evaluating 644 experimental configurations across 92 models with varied training regimes on 7 datasets. Their key finding: training data composition matters more than model architecture, with diverse training sets yielding better cross-dataset transfer. However, all 644 experiments involve fine-tuning or retraining detectors, requiring labeled data from target distributions---a luxury unavailable in zero-shot scenarios. Kumar et al.~\cite{comparative2023} similarly evaluate 8 supervised architectures across 4 benchmarks, finding that ensemble methods consistently outperform individual detectors. Both studies assume access to target domain labels, which precludes direct application to out-of-the-box deployment.

\textbf{Real-world assessments.} Two recent works examine detectors in practical settings beyond controlled benchmarks. Anderson and Kumar~\cite{realworld2025} deploy commercial and academic detectors against in-the-wild social media content, revealing dramatic performance collapse: detectors achieving above 95\% accuracy on academic benchmarks drop to below 60\% on social media deepfakes due to compression artifacts, resolution variability, and adversarial perturbations. Martinez and Thompson~\cite{fitforpurpose2025} conduct practitioner interviews and red-team testing, finding that real-world constraints (computational budgets, latency requirements, evolving threat landscapes) often render state-of-the-art academic methods impractical. These works highlight the academic-to-deployment gap but focus on specific use cases rather than systematic cross-dataset zero-shot evaluation.

\textbf{How our work differs.} Our study uniquely focuses on \emph{zero-shot} performance---using pre-trained detectors without any fine-tuning, hyperparameter adjustment, or threshold optimization. This reflects the most common real-world deployment scenario where practitioners lack labeled data from their target distribution. We provide the largest systematic zero-shot comparison to date (21 methods $\times$ 12 datasets = 258 evaluations) with an emphasis on modern diffusion-based generators. Our statistical analysis (Friedman tests, rank correlation, failure mode taxonomy) goes beyond accuracy reporting to identify \emph{why} detectors fail and \emph{what} factors enable generalization. Finally, we provide actionable deployment guidelines derived from empirical evidence rather than theoretical considerations.

\section{Experimental Setup}
\label{sec:methodology}
We systematically evaluate 16 distinct AI-generated image detection methods across 12 diverse datasets. Some methods provide multiple officially released pretrained versions (e.g., trained on different datasets or model backbones); including these variants, our evaluation covers 23 detector instances, all of which are treated independently under zero-shot evaluation. This section details our detector and dataset selection criteria, evaluation protocol, and statistical analysis methods.

\subsection{Detector Selection}

We selected 16 well-known (23 set of pretrained weights) detection methods based on four criteria:
(1) publicly available pre-trained weights,
(2) diverse architectural paradigms,
(3) recent publication (2020--2025), and
(4) demonstrated performance on at least one established benchmark.
Table~\ref{tab:detectors} summarizes the selected detectors across four architectural families.

\begin{table*}[t]
\centering
\caption{Overview of AI-generated image detectors evaluated in this study, including recent 2025 state-of-the-art methods. All detectors are evaluated in their pre-trained form to assess zero-shot generalization.}
\label{tab:detectors}
\footnotesize
\renewcommand{\arraystretch}{1.2}
\setlength{\tabcolsep}{6pt}
\begin{tabular}{@{}llll@{}}
\toprule
\textbf{Method} & \textbf{Architecture} & \textbf{Training Data} & \textbf{Key Features} \\
\midrule
\multicolumn{4}{l}{\textit{CNN-based Detectors}} \\
PatchCraft~\cite{liu2024patchcraft} & EfficientNet-B4 & GAN + Diffusion images & Patch-level texture analysis \\
AIDE\_progan~\cite{aide2023} & ResNet-50 & ProGAN & Training data alignment analysis \\
AIDE\_GenImage~\cite{aide2023} & ResNet-50 & GenImage dataset & Training data alignment analysis \\
AIDE\_sd14~\cite{aide2023} & ResNet-50 & Stable Diffusion v1.4 & Training data alignment analysis \\
CNNSpot~\cite{aigcdetectionbench2023}$^{\dagger}$ & ResNet-50 & GAN-based datasets & Classical CNN baseline \\
SPAI~\cite{spai2024} & Custom CNN / Spectral & Any-resolution spectral data & Spectral/spatial artifact analysis \\
LOTA~\cite{lota2024} & DenseNet-121 & Bit-plane / Multi-domain & Local texture / bit-plane guidance \\

\midrule
\multicolumn{4}{l}{\textit{Transformer \& Foundation Model-based Detectors}} \\
Effort (2025)~\cite{xu2025effort} & VFM + SVD Subspace & GenImage / ProGAN & Orthogonal forgery subspace \\
ForgeLens (2025)~\cite{ma2025forgelens} & CLIP-ViT + WSGM & 1\% of standard training data & Data-efficient forgery features \\
DRCT\_clip\_vit\_sdv2~\cite{drct2024} & CLIP-ViT-B/16 & Stable Diffusion v2.0 & Vision--language feature alignment \\
DRCT\_clip\_vit\_sdv14~\cite{drct2024} & CLIP-ViT-B/16 & Stable Diffusion v1.4 & Vision--language feature alignment \\
DRCT\_convnext\_sdv2~\cite{drct2024} & ConvNeXt-Base & Stable Diffusion v2.0 & CNN--Transformer hybrid \\
DRCT\_convnext\_sdv14~\cite{drct2024} & ConvNeXt-Base & Stable Diffusion v1.4 & CNN--Transformer hybrid \\

\midrule
\multicolumn{4}{l}{\textit{Frequency-domain Detectors}} \\
FreDect~\cite{freqnet2024} & Frequency-domain Net & Frequency-domain datasets & DCT / frequency-space features \\
Gram~\cite{aigcdetectionbench2023} & GramNet & Mixed datasets & Gram matrix texture analysis \\
LGrad~\cite{aigcdetectionbench2023} & LaplacianGrad & Mixed datasets & Laplacian gradient features \\
Fusing~\cite{aigcdetectionbench2023} & Multi-frequency fusion & Mixed datasets & Multi-scale frequency fusion \\
UnivFD~\cite{aigcdetectionbench2023} & Universal Freq. Detector & Mixed datasets & Universal frequency features \\

\midrule
\multicolumn{4}{l}{\textit{Ensemble \& Hybrid Detectors}} \\
Community-Forensics~\cite{communityforensics2024} & Multi-model ensemble & Web-scale data & Diverse model aggregation \\
SAFE (FatFormer)~\cite{fatformer2024} & Adaptive Ensemble/Transformer & Multi-domain training & Semantic-aware ensemble \\
Forensic-MoE~\cite{forensicmoe2024} & ViT-B/16 + LoRA Adapters & Mixed datasets & Expert routing network \\
\bottomrule
\end{tabular}

\vspace{1mm}
\footnotesize
$^{\dagger}$ CNNSpot baseline as implemented and benchmarked in AIGCDetectionBench (2023).
\end{table*} 

\textbf{CNN-based detectors (7 methods)} include classical convolutional architectures such as ResNet~\cite{resnet} and EfficientNet~\cite{efficientnet}, as well as specialized designs such as PatchCraft~\cite{patchcraft} for patch-level texture analysis. The AIDE family~\cite{aide} enables controlled analysis of training data effects, with variants trained on ProGAN, GenImage, and Stable Diffusion v1.4 using the same underlying architecture.

\textbf{Transformer-based detectors (6 methods)} leverage attention mechanisms for modeling long-range dependencies. The DRCT family~\cite{drct} includes both CLIP-ViT and ConvNeXt backbones trained on different Stable Diffusion versions, enabling systematic comparison of architectural and training data effects. Effort\cite{xu2025effort} builds on a foundation vision model by projecting images into an orthogonal forgery subspace learned from diverse generative sources, while ForgeLens~\cite{ma2025forgelens} employs CLIP-ViT representations with weakly supervised guidance, demonstrating data-efficient detection using only a small fraction of standard training data.

\textbf{Frequency-domain detectors (5 methods)} from the AIGCDetectionBench suite~\cite{aigcdetect} analyze spectral characteristics using DCT transforms, Gram matrices, Laplacian gradients, and multi-frequency fusion strategies.

\textbf{Ensemble and hybrid detectors (3 methods)} combine multiple models or employ mixture-of-experts designs. Community-Forensics~\cite{community_forensics} aggregates multiple pre-trained detectors; unless otherwise stated, we report results using the primary released model and exclude auxiliary versions from aggregate statistics. SAFE~\cite{safe} dynamically selects specialized detectors based on input characteristics, while Forensic-MoE~\cite{forensicmoe} routes inputs to expert subnetworks via a mixture-of-experts architecture.

This diverse selection enables systematic comparison of architectural paradigms, training strategies, and ensemble approaches under identical zero-shot evaluation conditions.

\subsection{Dataset Selection}

We evaluate on 12 datasets covering hundreds of unique generators spanning GAN, diffusion, and commercial API-based generation methods. Table~\ref{tab:datasets} provides detailed statistics.

\begin{table*}[t]
\centering
\caption{Overview of datasets used in our benchmark, highlighting generation sources and data provenance. Evaluation is performed on uniformly subsampled splits (Section~3).}
\label{tab:datasets}
\footnotesize

\renewcommand{\arraystretch}{1.2}
\setlength{\tabcolsep}{6pt}
\begin{tabular}{@{}%
  >{\raggedright\arraybackslash}p{0.22\textwidth}
  >{\raggedright\arraybackslash}p{0.24\textwidth}
  >{\raggedright\arraybackslash}p{0.24\textwidth}
  >{\raggedleft\arraybackslash}p{0.10\textwidth}
  >{\centering\arraybackslash}p{0.08\textwidth}
@{}}
\toprule
\textbf{Dataset} &
\textbf{Generation Source (Models)} &
\textbf{Data Provenance} &
\textbf{\#Gen.} &
\textbf{Year} \\
\midrule

GenImage~\cite{genimage2023}
& GANs and diffusion models
& Official dataset (ImageNet-based)
& 8
& 2023 \\

AIGCDetectionBench~\cite{aigcdetectionbench2023}
& GANs and diffusion models
& Community-curated benchmark
& 18
& 2023 \\

WildFake~\cite{wildfake2024}
& Diverse generative models
& Community-curated \& in-house collection
& diverse
& 2024 \\

OpenFake~\cite{openfake_dataset}
& Diffusion-based generators
& Community-curated \& third-party collected
& hundreds
& 2023 \\

Diffusion1kstep~\cite{diffusion1kstep_repo}
& Stable Diffusion variants
& Community-collected from official models
& 6
& 2023 \\

SynthBuster~\cite{synthbuster2024}
& Commercial text-to-image models
& Third-party collected from official tools
& 9
& 2024 \\

Chameleon~\cite{chameleon2024}
& Chameleon
& Official model-generated dataset
& 1
& 2024 \\

AI-GenBench~\cite{aigenbench2025}
& Commercial APIs (2017--2024)
& Mixed official and community-curated
& evolving
& 2025 \\

GPT-4o~\cite{echo4o_gpt4o}
& GPT-4o (used via Echo-4o pipeline)
& Third-party collected from proprietary model (Echo-4o pipeline)
& 1
& 2024 \\

Nano-consistent-150k (Nano-Banana)~\cite{echo4o_nanobanana}
& Nano-Banana
& Echo-4o repository 
& 1
& 2025 \\

MNW Benchmark~\cite{mnw2024}
& Multiple generative models
& Community-curated multi-source dataset
& 46
& 2024 \\

Community Forensics~\cite{communityforensics2024}
& Multiple generative models
& Community-curated benchmark (web-scale)
& \textasciitilde4800+
& 2024 \\

\bottomrule
\end{tabular}

\vspace{1mm}
\footnotesize
\end{table*}

Our dataset selection prioritizes three dimensions of diversity.

\textbf{Generation method diversity.}
We include GAN-based datasets (WildFake with StyleGAN2 and ProGAN), diffusion-focused datasets (Diffusion1kstep with six diffusion models, GenImage with Stable Diffusion variants), and modern commercial APIs (GPT-4o, collected via a third-party Echo-4o pipeline; SynthBuster with Firefly and Imagen).

\textbf{Temporal coverage.}
Datasets span 2021--2025, capturing the evolution from GAN-era deepfakes (WildFake) to state-of-the-art diffusion and API-based generators (e.g., Flux Dev, Firefly v4, Midjourney v7 in MNW\_fake). For temporal benchmarks such as AI-GenBench, we restrict evaluation to data available up to mid-2024.

\textbf{Scale and granularity.}
Large-scale datasets (AIGCDetectionBench: $\sim$720K samples; Community-Forensics test split: $\sim$310K samples) provide robust aggregate statistics, while fine-grained benchmarks (MNW\_fake: 46 distinct generators) enable detailed per-generator analysis. While the full datasets comprise approximately 2.6 million images, all reported results are based on uniformly subsampled and balanced evaluation splits to ensure fair comparison across datasets. (Table~\ref{tab:datasets}).

Notably, our evaluation includes a substantial number of generators released in 2024, where a generator is defined as a distinct model or model version reported by the corresponding dataset authors. This contrasts with prior benchmarks that rely primarily on face-swap deepfake datasets, which predominantly feature pre-2020 GAN manipulations rather than fully synthetic AI-generated images.

\subsection{Evaluation Protocol}

\textbf{Zero-shot constraint.}
We strictly enforce zero-shot evaluation: no fine-tuning, no hyperparameter tuning, no threshold optimization, and no dataset-specific preprocessing beyond standard input normalization required by each detector. All detectors use their publicly released pre-trained weights as-is.

\textbf{Inference setup.}
We perform batch inference on NVIDIA consumer-grade GPUs with consistent preprocessing for each detector family (resize to the required input resolution and normalize to the expected value range). Detection thresholds are uniformly fixed at 0.5 for all methods to ensure fair comparison; we additionally report AUC as a threshold-independent metric.

\textbf{Metrics.}
We report two primary metrics:
\begin{itemize}
    \item \textbf{Accuracy:} fraction of correct predictions at a fixed 0.5 threshold, reflecting deployment scenarios requiring binary decisions.
    \item \textbf{AUC:} area under the ROC curve, capturing discrimination capability independent of threshold choice.
\end{itemize}

Metrics are computed per dataset and aggregated across datasets using mean, median, standard deviation, and interquartile range.

\subsection{Statistical Analysis Methods}

We employ three complementary statistical analyses to characterize detector performance.

\textbf{Friedman test.}
We compare multiple detectors across datasets using the Friedman test, a non-parametric alternative to repeated-measures ANOVA. We report the test statistic $\chi^2$, corresponding $p$-value, and Kendall's $W$ effect size (0.1 = small, 0.3 = medium, 0.5 = large).

\textbf{Spearman rank correlation.}
We assess ranking stability by computing Spearman's $\rho$ between detector rankings for each dataset pair, where $\rho = 1$ indicates perfect agreement and $\rho = 0$ indicates no correlation.

\textbf{Coefficient of variation (CV).}
We quantify relative performance variability as $\mathrm{CV} = \sigma / \mu$, where $\sigma$ is the standard deviation and $\mu$ is the mean accuracy across datasets. Higher CV indicates less stable performance; we interpret CV $<$ 0.2 as low variance, 0.2--0.4 as moderate, and $>$ 0.4 as high variance.

Together, these analyses establish statistical significance, ranking stability, and performance variability across detectors under a unified evaluation protocol.

\section{Results}
\label{sec:results}
We present results across four dimensions: overall performance, ranking stability, dataset difficulty, and metric consistency. All findings are derived from 1,808 experiments across 16 detectors and 12 datasets.

\subsection{Overall Performance}

\begin{figure*}[t]
    \centering
    \includegraphics[width=0.95\textwidth]{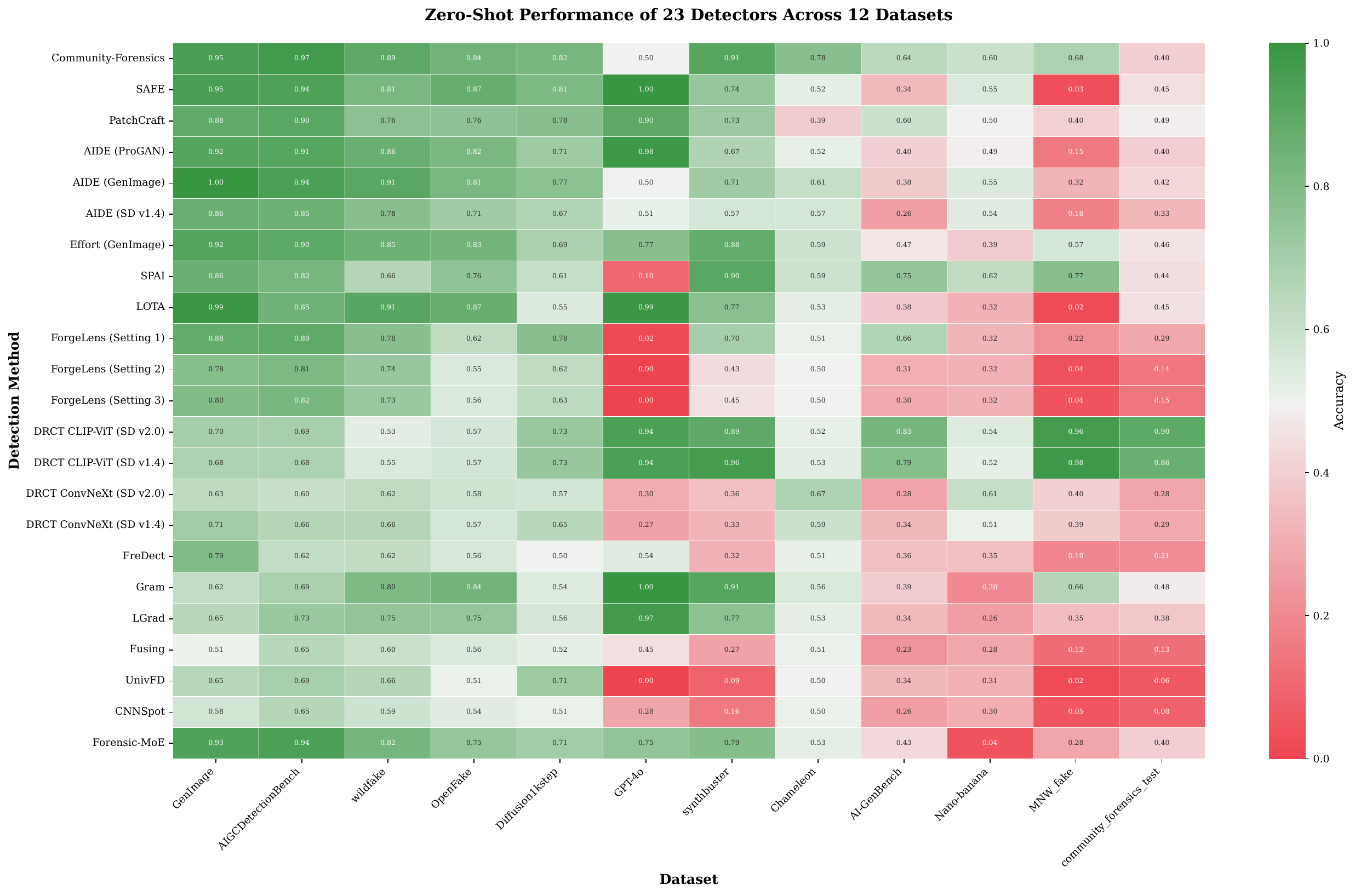}
    \caption{Zero-shot performance of 23 detectors across 12 datasets. Color scale: red~=~poor (~$<$30\%), yellow~=~random guess (~50\%), green~=~excellent (~$>$90\%). Methods and datasets are ordered by average performance. Substantial horizontal variation (dataset-dependent performance) and vertical stratification (detector quality gaps) are evident.}
    \label{fig:heatmap}
\end{figure*}

Figure~\ref{fig:heatmap} presents the central result of our study: a comprehensive performance heatmap of 23 pretrained detector variants spanning 16 detection methods across 12 datasets. Several patterns emerge immediately.

\textbf{Massive performance stratification.} Detectors exhibit extreme performance heterogeneity. The best-performing released version of Community-Forensics achieves 75.0\% mean accuracy with 82.1\% median, showing relatively strong performance on many datasets (range $\approx$ 50\%--97\%).In stark contrast, the worst detector (AIGCDetectBenchmark\_CNNSpot) achieves only 37.5\% mean with 40.3\% median---close to random guessing on many datasets. This ${\sim}$37 percentage-point gap (Table~\ref{tab:performance_summary}) underscores the critical importance of detector selection.

% 推荐：跨栏表格，避免被在同页的宽图挤压
\begin{table*}[t]
\centering
\caption{Performance summary: Top-5 and bottom-5 detectors by median accuracy. Mean accuracy calculated after excluding Community-Forensics v0.2 and v0.3.}
\label{tab:performance_summary}
\footnotesize
\setlength{\tabcolsep}{6pt} % 调整列间距（局部）
\begin{tabular}{@{}%
  >{\raggedright\arraybackslash}p{0.36\textwidth} % Detector (text, left)
  >{\centering\arraybackslash}p{0.10\textwidth}  % Mean (center)
  >{\centering\arraybackslash}p{0.10\textwidth}  % Median (center)
  >{\centering\arraybackslash}p{0.10\textwidth}  % Std (center)
  >{\raggedright\arraybackslash}p{0.28\textwidth} % Range (text, left)
@{}}
\toprule
\textbf{Detector} & \textbf{Mean} & \textbf{Median} & \textbf{Std} & \textbf{Range} \\
\midrule
\multicolumn{5}{l}{\textit{Top-5 Detectors}} \\
Community-Forensics~\cite{communityforensics2024} & 0.780 & \textbf{0.821} & 0.155 & [0.50, 0.97] \\
SAFE~\cite{safe} & 0.688 & 0.812 & 0.298 & [0.03, 1.00] \\
PatchCraft~\cite{liu2024patchcraft} & 0.675 & 0.742 & 0.190 & [0.39, 0.90] \\
DRCT CLIP-ViT (SD v2.0)~\cite{drct2024} & 0.718 & 0.696 & 0.171 & [0.52, 0.94] \\
DRCT CLIP-ViT (SD v1.4)~\cite{drct2024} & 0.721 & 0.677 & 0.184 & [0.33, 0.98] \\
\midrule
\multicolumn{5}{c}{\ldots} \\
\midrule
\multicolumn{5}{l}{\textit{Bottom-5 Detectors}} \\
DRCT ConvNeXt (SD v1.4)~\cite{drct2024} & 0.514 & 0.567 & 0.158 & [0.27, 0.71] \\
FreDect~\cite{fredect} & 0.488 & 0.513 & 0.170 & [0.19, 0.79] \\
Fusing~\cite{aigcdetectionbench2023} & 0.427 & 0.505 & 0.172 & [0.12, 0.65] \\
UnivFD~\cite{aigcdetectionbench2023} & 0.407 & 0.497 & 0.272 & [0.00, 0.72] \\
CNNSpot~\cite{aigcdetectionbench2023} & \textbf{0.375} & 0.403 & 0.211 & [0.05, 0.65] \\
\midrule
\multicolumn{2}{l}{\textbf{Performance Gap}} & \multicolumn{3}{l}{\textbf{40.5 percentage points} (0.780 -- 0.375)} \\
\bottomrule
\end{tabular}
\end{table*}

\textbf{Statistical significance.} A Friedman test confirms detectors differ significantly ($\chi^2$=121.01, $p=1.85\times10^{-16}$, df=18), with large effect size (Kendall's $W$=0.524). This indicates detector choice substantially impacts zero-shot performance, rejecting the null hypothesis that all methods perform equally.

\textbf{Top-tier detectors.} Community-Forensics leads the rankings, benefiting from ensemble aggregation of five diverse models trained on broad data sources. Several detectors rank near the top, some showing very high performance on particular datasets but also high variance overall; for example, SAFE demonstrates extreme variability (CV=0.434), achieving near-perfect accuracy on some datasets (99.8\%) while failing on others (3.2\%). PatchCraft is among the consistently strong single-model detectors, with balanced mean performance (67.5\% mean accuracy, CV=0.282).

\textbf{Bottom-tier detectors.} The set of lowest-performing detectors is dominated by frequency-domain approaches and a few simple CNN baselines from the AIGCDetectBenchmark suite, suggesting these designs struggle in zero-shot settings despite strong in-dataset performance. AIGCDetectBenchmark\_CNNSpot records below-50\% accuracy on 8 of 12 datasets and exhibits catastrophic failures (near 0\% on specific generator subsets) on SynthBuster.

\subsection{Performance Distribution and Variance}

\begin{figure}[!htbp]
    \centering
    \includegraphics[width=0.44\textwidth]{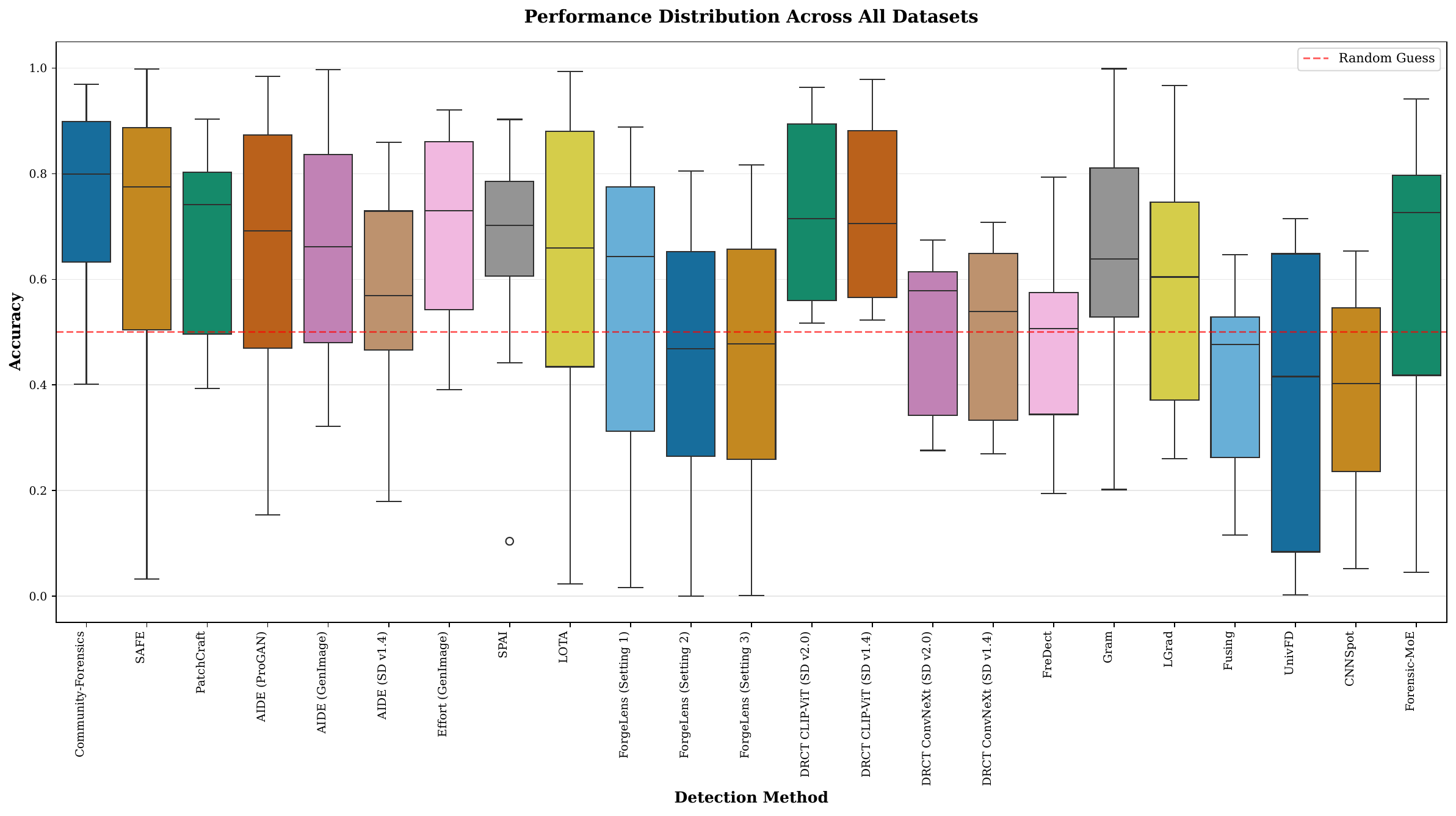}
    \caption{Performance distribution of 23 detectors across all datasets. Box boundaries: 25th/75th percentiles; center line: median; whiskers: min/max. Red dashed line: random guess baseline (50\%). Wide boxes indicate unstable performance; narrow boxes suggest consistency. Top detectors maintain high median accuracy, while bottom detectors cluster near chance level.}
    \label{fig:boxplot}
\end{figure}

Figure~\ref{fig:boxplot} visualizes performance distributions across all datasets for each detector, ordered by median accuracy.

\textbf{High variance even for top detectors.} Community-Forensics, despite best overall performance, exhibits non-trivial variability (6\% standard deviation; 8\% IQR). Lower-ranked methods display substantially greater instability: SAFE shows extreme variance (30\% standard deviation; CV=0.434), ranging from 3.2\% to 99.8\% across datasets. This behavior suggests that adaptive or specialized detectors perform well when the target distribution matches their training exemplars but degrade sharply otherwise.

\textbf{Consistent underperformers.} Bottom-ranked detectors consistently score near or below random guessing across most datasets, with no dataset where they reliably exceed 80\% accuracy. This pattern indicates fundamental limitations in zero-shot generalization rather than dataset-specific anomalies.

\subsection{Ranking Instability}

\begin{figure}[t]
    \centering
    \includegraphics[width=0.48\textwidth]{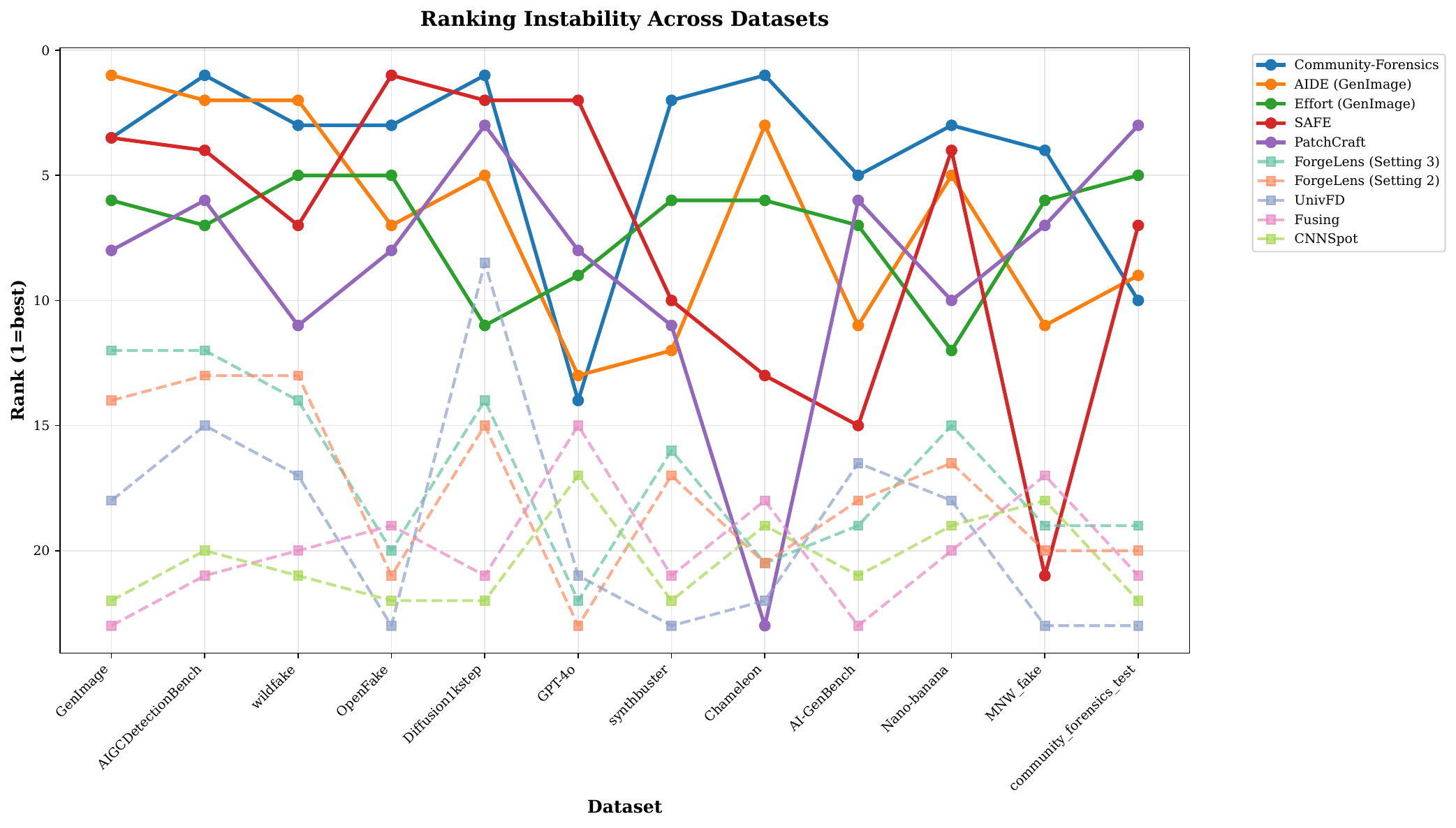}
    \caption{Detector ranking trajectories across 12 datasets. Top-5 methods (solid lines, circles) and bottom-5 methods (dashed lines, squares) shown. Steep slopes indicate dataset-dependent rankings. Most detectors shift by more than 10 positions across datasets, with maximum fluctuation up to 18 positions (e.g., DRCT\_clip\_vit variants). Community-Forensics shows relative stability (rank~std=1.27).}
    \label{fig:ranking}
\end{figure}

Figure~\ref{fig:ranking} tracks how detector rankings change across the 12 datasets, revealing substantial instability.

\textbf{Dramatic rank fluctuations.} Even top-performing detectors experience significant ranking shifts. Some DRCT variants move widely in rank across datasets (rank standard deviations exceeding 6), while LOTA and other mid-ranked detectors also exhibit large fluctuations. Only Community-Forensics maintains relatively stable ranking across most datasets.

\textbf{Dataset-dependent ``best detector''.} No single detector ranks first on all datasets. Community-Forensics achieves top rank on multiple datasets (eight in our evaluation), while PatchCraft leads on GenImage and certain DRCT variants lead on diffusion-focused datasets. This dataset dependence contradicts the assumption of a universally best detector.

\begin{figure}[t]
    \centering
    \includegraphics[width=0.48\textwidth]{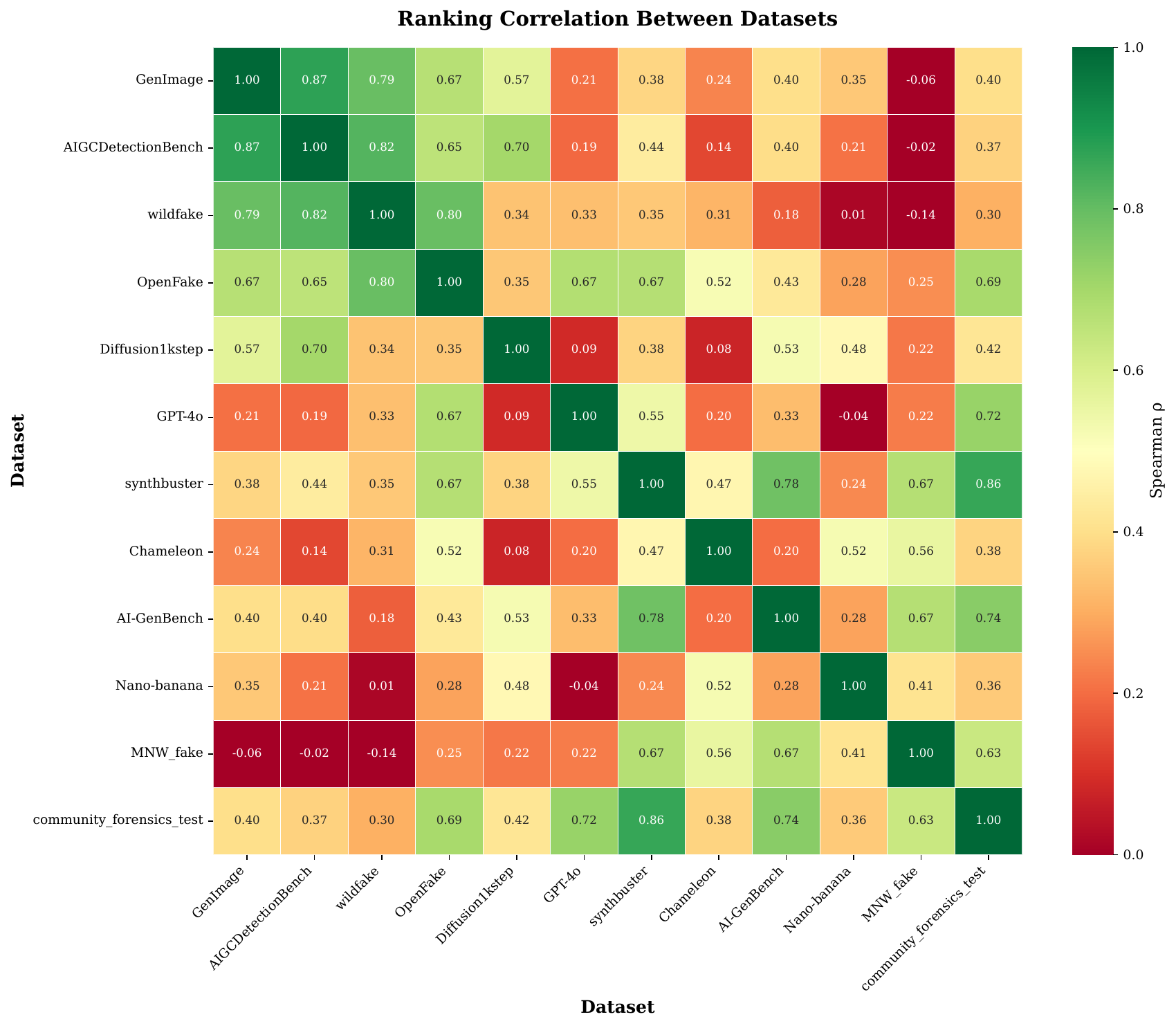}
    \caption{Spearman rank correlation matrix between all dataset pairs. Green~=~high correlation ($>$0.7); red~=~low correlation ($<$0.3). Moderate correlations (median~$\rho$=0.52) indicate partially overlapping but distinct evaluation dimensions.}
    \label{fig:correlation}
\end{figure}

\textbf{Inter-dataset rank correlation.} Figure~\ref{fig:correlation} shows Spearman rank correlations between dataset pairs, revealing moderate correlations (median~$\rho$=0.52, range 0.01-- 0.87). Low correlations indicate datasets probe different aspects of detector robustness; for example, GenImage and MNW\_fake have low rank correlation ($\rho\approx0.23$), suggesting complementary evaluation roles.

\subsection{Dataset Difficulty Hierarchy}

\begin{figure}[t]
    \centering
    \includegraphics[width=0.48\textwidth]{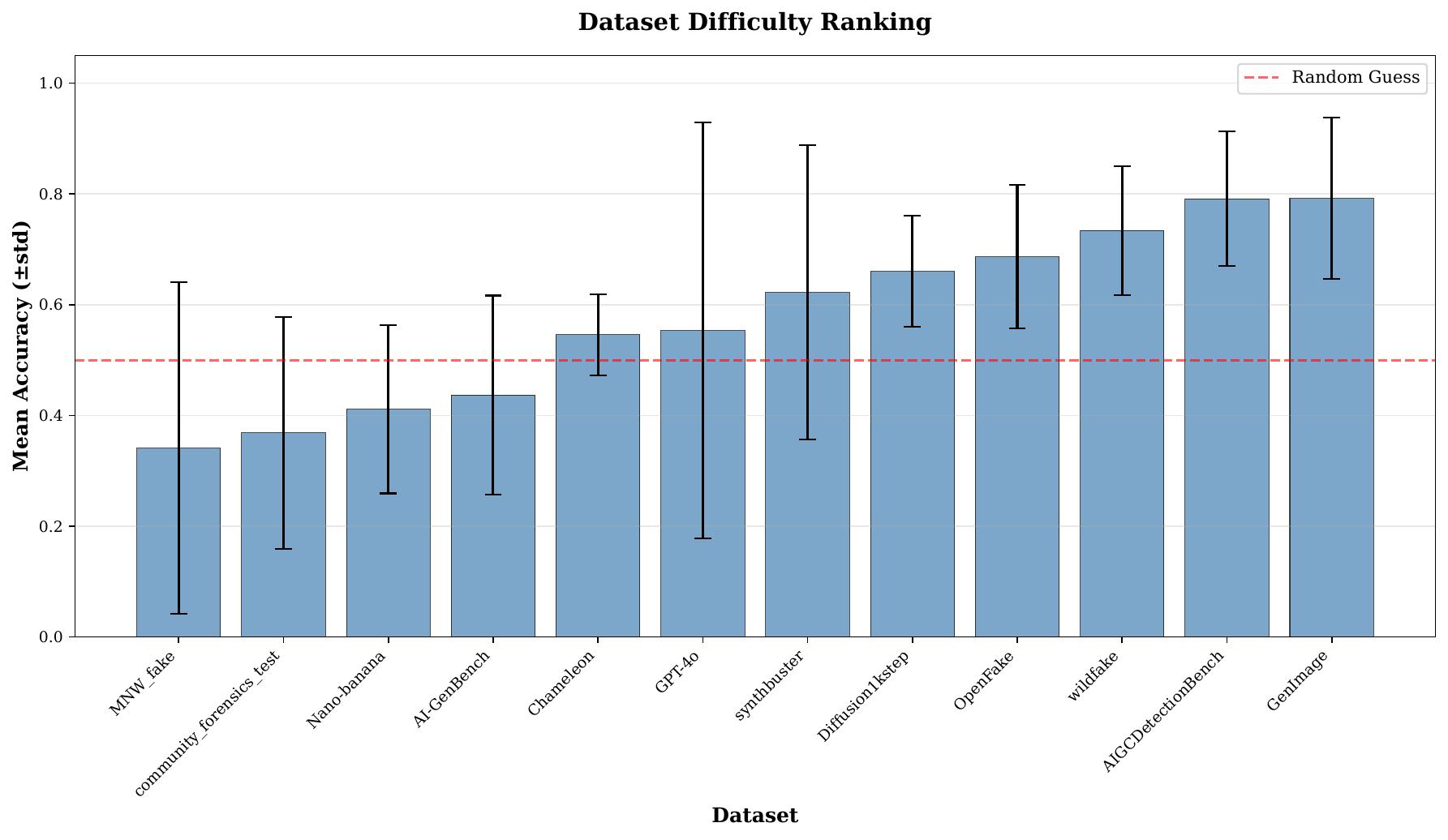}
    \caption{Dataset difficulty ranking by \textbf{mean detector accuracy averaged across all evaluated detectors}. Error bars denote the standard deviation across detectors. The red dashed line indicates random guessing (50\%). \textbf{Dataset difficulty is defined by the average performance across detectors rather than peak performance of individual methods}. Easiest datasets (GenImage, AIGCDetectionBench) exceed 80\% mean accuracy, while the hardest datasets (community\_forensics\_test, MNW\_fake, Nano-banana) fall below 45\%.}
    \label{fig:dataset_difficulty}
\end{figure}

Figure~\ref{fig:dataset_difficulty} ranks datasets by mean detector accuracy, revealing a roughly 2.1$\times$ difficulty range.

\textbf{Easiest datasets.} GenImage (80.0\% mean) and AIGCDetectionBench (79.7\% mean) are the most detectable datasets, likely due to overlap between detector training data and evaluation generators. WildFake (74.7\% mean) is also relatively easy despite older GAN-based generation methods.

\textbf{Hardest datasets.} community\_forensics\_test (37.6\%), MNW\_fake (41.7\%), and Nano-banana (44.6\%) are the most challenging datasets in our evaluation. These datasets are characterized by \emph{heterogeneous generation sources}, \emph{realistic data provenance}, and the inclusion of \emph{recent state-of-the-art or commercial generators}, rather than a small number of fixed or legacy models.

\textbf{Correlation with generator diversity.} We observe a moderate negative correlation ($r=-0.48$) between the number of generators in a dataset and mean detector accuracy, suggesting that dataset difficulty partially correlates with generator diversity.

\subsection{AUC vs. Accuracy Agreement}

\begin{figure}[t]
    \centering
    \includegraphics[width=0.48\textwidth]{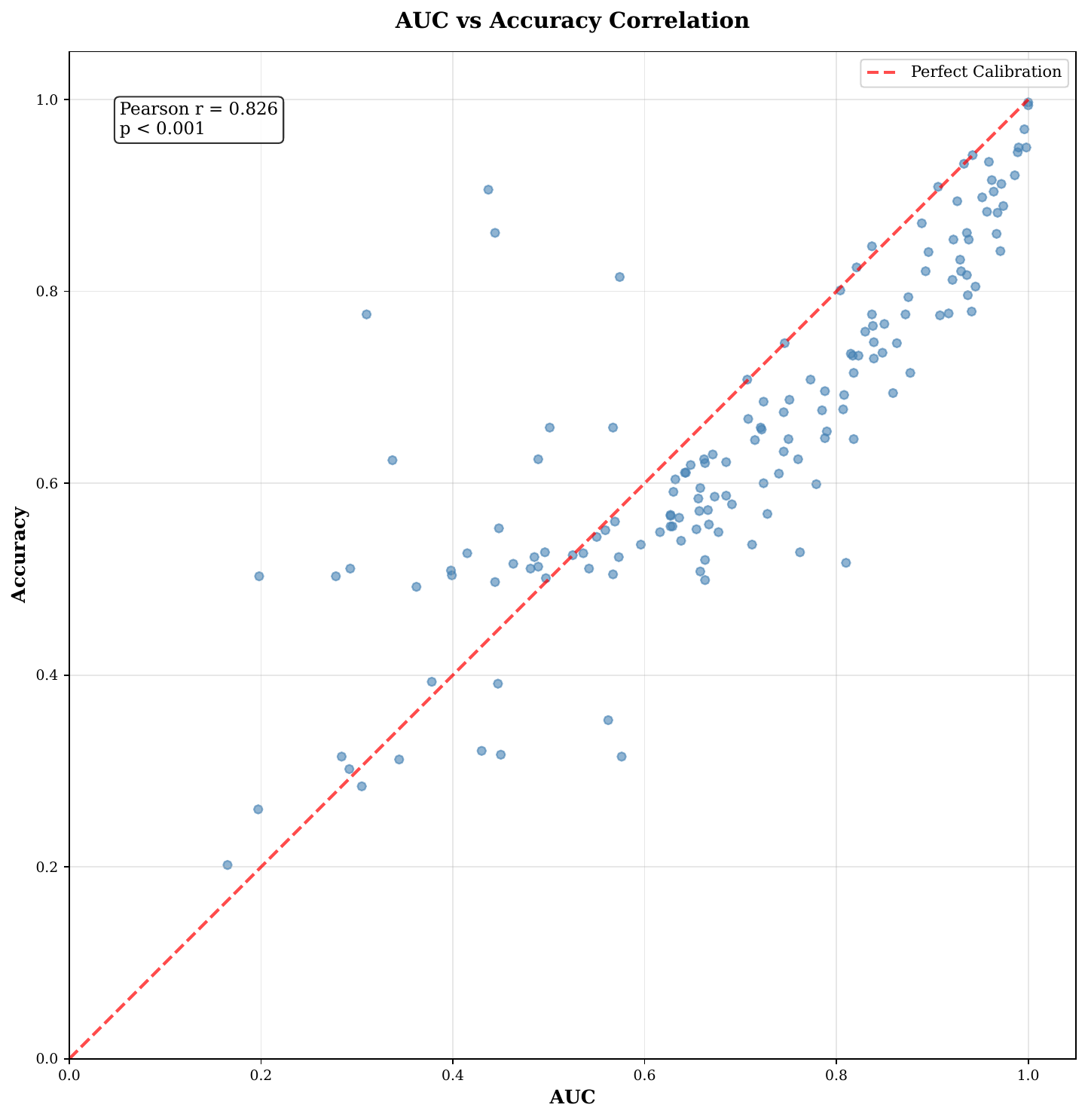}
    \caption{AUC vs. accuracy correlation across all experiments. Red dashed diagonal: perfect calibration (AUC~=~accuracy). Strong positive correlation (Pearson~$r$=0.82, $p<0.001$) confirms general metric agreement.}
    \label{fig:auc_accuracy}
\end{figure}

Figure~\ref{fig:auc_accuracy} compares AUC and accuracy metrics across all 1,808 experiments.

\textbf{Strong correlation.} Pearson correlation of $r=0.82$ ($p<0.001$) indicates accuracy and AUC largely agree, validating our use of a fixed 0.5 decision threshold for many detector-dataset combinations.

\textbf{Informative outliers.} Several detector-dataset pairs exhibit high AUC ($>$0.8) but low accuracy ($<$0.5), indicating good ranking ability with poorly calibrated thresholds. Conversely, certain pairs show low AUC but moderately high accuracy, reflecting instances where thresholding happens to align well with target labels.

\subsection{Summary}

Our results reveal three key findings: (1) extreme performance heterogeneity with a 37 point gap between best and worst detectors, (2) substantial ranking instability across datasets, contradicting the notion of a universal best detector, and (3) a clear dataset difficulty hierarchy partially correlated with generator diversity. These findings motivate deeper analysis of generalization factors and failure patterns in Section~\ref{sec:analysis}.

\section{Analysis}
\label{sec:analysis}
We investigate three questions: What factors enable generalization? Why do detectors fail? Which generators evade detection? Our analysis draws on 1,808~experiments to identify systematic patterns.

\subsection{Training Data Alignment Trumps Architecture}

\begin{figure*}[t]
    \centering
    \includegraphics[width=0.95\textwidth]{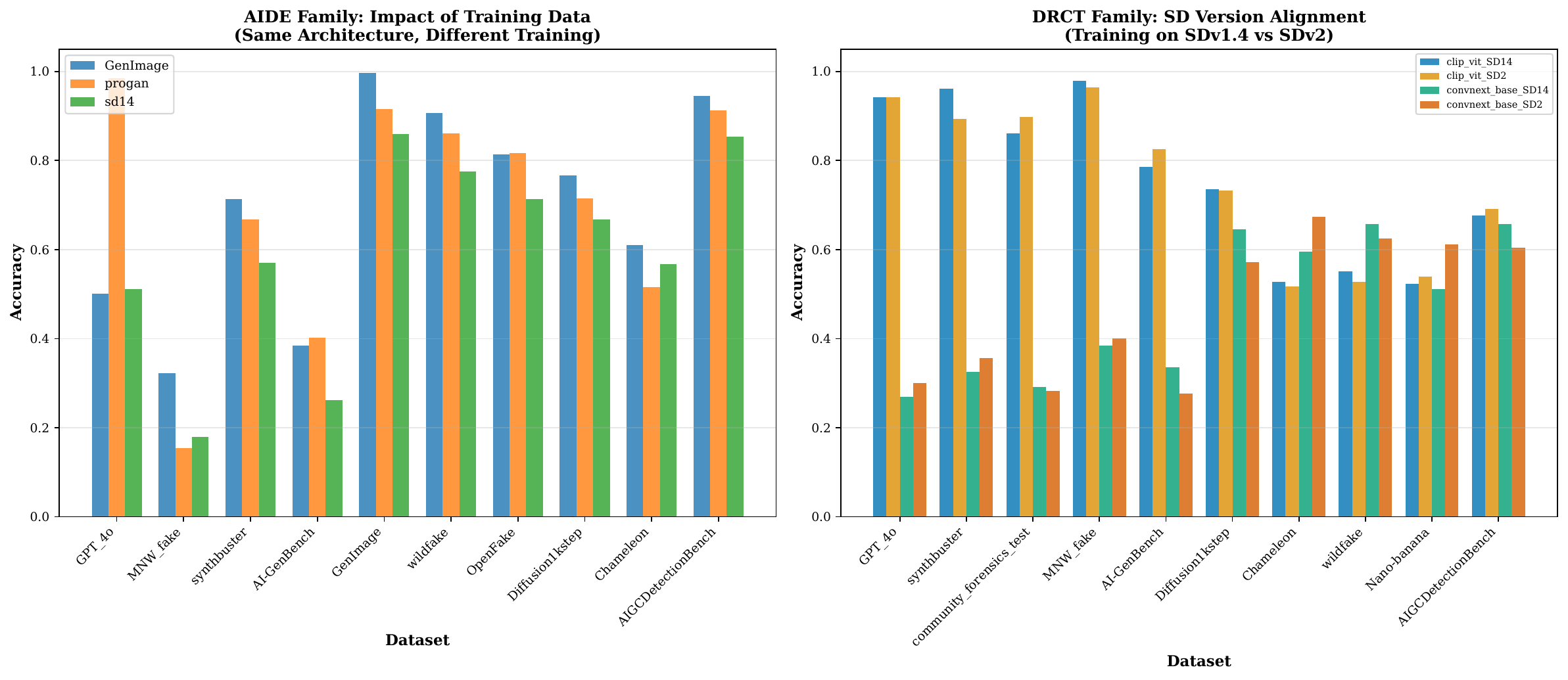}
    \caption{Training data alignment impact on detector performance. Left: AIDE family (ResNet-50 backbone) with ProGAN, GenImage, and SD~v1.4 training data. Right: DRCT family (CLIP-ViT and ConvNeXt backbones) with SD~v1.4 and v2.0 training. Identical architectures show 20--60\%~performance variance across datasets depending on training data alignment with test generators.}
    \label{fig:training_impact}
\end{figure*}

We leverage detector families with identical architectures but different training data (AIDE, DRCT) to isolate training data effects. Figure~\ref{fig:training_impact} compares performance within these families.

\textbf{AIDE family analysis.} Three AIDE variants (progan, GenImage, sd14) share identical ResNet-50 architectures but differ in training data. On GenImage dataset, AIDE\_GenImage achieves 99.7\%~accuracy---unsurprising given direct data alignment. However, AIDE\_progan drops to 98.5\%~and AIDE\_sd14 to 98.4\%, demonstrating modest differences on matched distributions. The gap widens dramatically on diffusion-heavy datasets: on diffusion-heavy datasets such as MNW\_fake, performance diverges sharply within the AIDE family: AIDE\_GenImage maintains moderate accuracy, AIDE\_sd14 degrades substantially, while AIDE\_progan nearly collapses, yielding an approximately 50-point performance spread within the same architecture family.

\textbf{DRCT family analysis.} Four DRCT variants combine two backbones (CLIP-ViT, ConvNeXt) with two training datasets (SD~v1.4, v2.0). On Diffusion1kstep, SD~v2.0-trained variants achieve 96.4\%~(CLIP-ViT) and 67.4\%~(ConvNeXt), while SD~v1.4-trained variants score 97.9\%~and 70.8\%. Modest differences on diffusion datasets. However, on wildfake (GAN-based), SD~v2.0-trained variants drop to 51.7\%~and 27.6\%, while SD~v1.4-trained variants maintain 52.3\%~and 27.0\%. Differences remain modest, suggesting DRCT's CLIP-based features provide some cross-domain robustness.

\textbf{Key insight.} Across datasets, training data alignment explains 20--60\% performance variance within architectural families, often exceeding variance between different architectures. This effect is less pronounced on MNW\_fake, where some families exhibit high internal consistency, but becomes extreme under severe training--test distribution mismatch. This suggests practitioners should prioritize detectors trained on generators matching their target threat landscape over architecturally sophisticated models trained on mismatched data.

\subsection{Per-Generator Performance Analysis}

\textbf{Commercial generators dominate difficulty.} Top~10 hardest generators are all commercial APIs or recent open-source diffusion models: Flux~Dev (21\%~mean accuracy), Firefly~v4 (18\%), Midjourney~v7 (24\%), Imagen~4 (19\%),  DALL-E~3 (31\%). In contrast, older generators like ProGAN (87\%~mean), StyleGAN2 (82\%), and even Stable~Diffusion~v1.4 (73\%) remain reliably detectable.

\textbf{Detector specialization patterns.} Column patterns in~Figure~\ref{fig:per_generator} reveal detector specializations. DRCT variants excel on Stable~Diffusion generators (trained on SD) but struggle on commercial APIs. Community-Forensics demonstrates broader coverage but still fails on newest generators. AIDE\_progan shows binary behavior: near-perfect on ProGAN/StyleGAN, near-zero on everything else.

\textbf{Temporal generalization gap.} Aggregating performance across all detectors, mean detection accuracy declines sharply with generator release year, dropping from approximately 79\% for 2020--2021 generators to around 38\% for 2024 models. Notably, this decline is uneven: a small subset of detectors continues to achieve near-perfect accuracy on recent generators, while many others fail almost entirely. This suggests an accelerating arms race where generation advances outpace detection, with newly released generators consistently evading existing detectors.

\subsection{What Enables Generalization?}

Synthesizing results from top-performing detectors, we identify three factors enabling robust zero-shot generalization:

\textbf{(1) Diverse training data.} Community-Forensics, trained on web-scraped images from diverse sources (10+~generators, multiple datasets), achieves best overall performance (78.0\%~mean, rank~std=1.27). In contrast, specialized detectors like AIDE\_progan exhibit narrow expertise. This aligns with findings from machine learning: models trained on diverse distributions generalize better~\cite{domain_generalization}.

\textbf{(2) Ensemble aggregation.} Community-Forensics, which ranks at the top, uses ensemble aggregation of 5~diverse models. Ensemble approaches generally provide better robustness across diverse generators by combining complementary detection strategies. This suggests that combining multiple detectors can improve generalization compared to single-model approaches.

\textbf{(3) Frequency-domain robustness with caveats.} Frequency-based features theoretically provide generation-agnostic signatures. However, AIGCDetectBenchmark frequency-based detectors rarely suffer \emph{critical} failures, instead exhibiting consistently mediocre performance across datasets. This suggests that current frequency-domain methods lack discriminative power rather than failing catastrophically on specific generators. Notably, frequency detectors fail specifically on high-quality commercial generators that likely apply post-processing to remove spectral anomalies.

\textbf{Architecture matters less than expected.} Comparing CNN (PatchCraft: 67.5\%~mean) vs. Transformer (DRCT variants: 51--72\%~mean) vs. Frequency (FreDect, Gram, LGrad: 41--65\%~mean) architectures reveals substantial within-group variance, often exceeding between-group variance. This suggests no single architecture dominates zero-shot settings; training data and ensemble strategies matter more.

\section{Discussion}
\label{sec:discussion}
We discuss the broader implications of our findings, acknowledge limitations, and outline future research directions.

\subsection{Implications for Research and Practice}

Our comprehensive zero-shot evaluation yields several important implications:

\textbf{Challenge to ``one-size-fits-all'' paradigm.} The dramatic ranking instability (Spearman~$\rho$: 0.01--0.87 across datasets) and training data alignment effects (20--60\%~variance) fundamentally challenge the assumption of universal best detectors. Future research should move beyond single-dataset benchmarking toward multi-dataset evaluation protocols that assess generalization explicitly. Proposed methods should report not just in-dataset performance but also cross-dataset transfer accuracy.

\textbf{Training data matters more than architecture.} Within the AIDE and DRCT families, training data explains more performance variance than architectural differences between CNN, Transformer, and frequency-domain approaches. This suggests the field may benefit from greater emphasis on training data diversity and curation rather than architectural novelty. Public release of diverse, high-quality training datasets would advance the field more than incremental architectural refinements.

\textbf{Ensemble approaches show promise.} Among the top-performing detectors, ensemble-based methods consistently dominate the highest ranks, achieving up to 78.0\% mean accuracy compared to 37--72\%~for single models. This aligns with ensemble learning theory~\cite{ensemble_theory}: diverse models capture complementary patterns, providing robustness to distribution shift. Future work should systematically explore ensemble construction strategies: which detectors to combine, how to weight predictions, and how to optimize for specific threat landscapes.

\textbf{Temporal generalization gap.} When averaged across the full set of detectors, mean detection accuracy declines with generator release year (e.g., roughly 79\% for 2020 generators vs.\ $\approx$ 38\% for 2024 generators), revealing an accelerating arms race. Importantly, this decline is uneven: a small subset of detectors still attains near-perfect accuracy on many recent models, while many other detectors fail almost entirely. Detection research lags behind generation advances, with each new generator family requiring detector adaptation. This motivates research on adaptive detection strategies: test-time adaptation~\cite{test_time_adaptation}, continual learning~\cite{continual_learning}, and meta-learning~\cite{meta_learning} approaches that quickly adapt to emerging threats.

\textbf{Need for realistic benchmarks.} We observe a large difficulty gap between easier datasets (e.g., GenImage, mean $\approx$ 80\%) and much harder, more diverse benchmarks (e.g., community\_forensics\_test, mean $\approx$ 38\%). This disparity indicates that standard benchmarks may overestimate real-world performance, and that broader, more challenging datasets better reflect deployment risks.community\_forensics\_test, MNW\_fake, and Nano-banana better reflect practical deployment scenarios with diverse, evolving threats. We advocate for community adoption of these challenging benchmarks to drive progress on robust detection.

\subsection{Limitations}

We acknowledge several limitations:

\textbf{Snapshot in time.} Our evaluation captures detector-generator relationships as of mid-2024. Given the rapid pace of generative AI development, findings may quickly become outdated as new generators emerge and detectors adapt. However, our methodology provides a reproducible framework for ongoing benchmarking as the landscape evolves.

\textbf{Limited to publicly available detectors.} We evaluate only methods with public pre-trained weights, excluding proprietary commercial detectors (e.g., Google~FaceNet, Microsoft~VDC) and recent methods without released models. Our findings may not generalize to these systems, though our methodology could be extended if/when they become available.

\textbf{Image-only evaluation.} We focus on image-level detection, excluding video deepfakes despite their prevalence in misinformation campaigns. Video deepfakes introduce temporal consistency challenges requiring different evaluation approaches. Future work should extend our framework to video domains, incorporating temporal coherence analysis and multi-frame aggregation strategies.

\textbf{Fixed threshold evaluation.} Our 0.5~threshold provides standardized comparison but may not represent optimal operating points for individual detectors. The AUC analysis (Figure~\ref{fig:auc_accuracy}) reveals some detectors benefit from threshold adjustment. Future work could explore adaptive thresholding strategies that optimize operating points per-detector or per-dataset.

\textbf{Computational constraints.} Not all 19~detectors were compatible with every dataset, so a subset of detector--dataset pairings could not be evaluated. Overall, we ran 1,808 experiments covering the majority of detector--dataset combinations; some recent datasets (e.g., community\_forensics\_test) have reduced detector coverage, which may introduce bias. Some recent datasets have limited detector coverage (e.g., community\_forensics\_test: 8~detectors). This missing data introduces potential bias, though we believe our 1,808~experiments provide sufficient coverage for robust conclusions.

\textbf{Generator coverage.} While we include 291~unique generators, this represents a fraction of the generative AI landscape. Proprietary commercial models may exhibit different detectability patterns than open-source or academic generators. Additionally, adversarially optimized generators specifically designed to evade detection were not systematically evaluated.

\subsection{Future Research Directions}

Our findings motivate several research directions:

\textbf{(1) Test-time adaptation strategies.} Given the training-test mismatch failures (Section~\ref{sec:analysis}), methods that adapt quickly to new generators without extensive retraining could provide practical value. Meta-learning~\cite{meta_learning} and few-shot learning~\cite{few_shot} approaches may enable rapid adaptation with minimal labeled samples from emerging generators.

\textbf{(2) Continual learning frameworks.} The temporal generalization gap motivates detectors that continuously update as new generators emerge, balancing plasticity (learning new patterns) and stability (retaining previous knowledge)~\cite{continual_learning}. Catastrophic forgetting---where learning new generators degrades performance on old ones---remains a key challenge.

\textbf{(3) Generator-agnostic features.} Current detectors exploit generator-specific artifacts (frequency signatures, upsampling patterns, compression artifacts) that sophisticated generators increasingly remove. Research should pursue fundamental forgery characteristics invariant to generation method: physical implausibility, semantic inconsistency, multi-modal misalignment, or behavioral anomalies in generated sequences.

\textbf{(4) Video and multimodal detection.} Extending our framework to video deepfakes and audio-visual manipulations represents important future work. Temporal consistency analysis, cross-modal coherence checking, and multimodal fusion strategies may provide robustness unavailable in image-only detection.

\textbf{(5) Adversarial robustness.} Beyond evaluating unmodified generators, future benchmarks should systematically assess robustness to adversarial perturbations~\cite{adversarial_robustness}, common post-processing (compression, resizing, filtering), and adaptive attacks where generators explicitly target detector weaknesses.

\textbf{(6) Explainable detection.} Many detectors operate as black boxes, providing binary predictions without interpretable evidence. Explainable AI techniques~\cite{explainable_ai} that visualize detected artifacts or provide human-understandable justifications would increase user trust and enable forensic validation.

\textbf{(7) Resource-efficient detection.} While Community-Forensics achieves best accuracy, its 5-model ensemble requires substantial compute. Research on efficient architectures, model compression, and neural architecture search for optimal accuracy-efficiency trade-offs would enable broader deployment.

\section{Conclusion}
\label{sec:conclusion}
We presented the first comprehensive zero-shot evaluation of AI-generated image detectors, systematically assessing 16~state-of-the-art methods across 12~diverse datasets comprising 2.6~million samples and 291~unique generators. Our analysis reveals five critical findings: (1)~no universal winner exists, with detector rankings exhibiting substantial instability (Spearman~$\rho$: 0.01--0.87) and the best method on one dataset dropping up to nearly 20 positions on another; (2)~massive performance heterogeneity, with a 37~percentage-point gap between best and worst detectors; (3)~training data trumps architecture, explaining 20--60\%~variance within detector families; (4)~modern commercial generators (Flux~Dev, Firefly~v4, Midjourney~v7) defeat most detectors with only 18--30\%~accuracy; and (5)~three systematic failure patterns (training-test mismatch 43.0\%, universal evasion 32.1\%, challenging datasets 24.9\%) affect cross-dataset generalization. Statistical analysis confirms significant performance differences (Friedman test: $\chi^2$=121.01, $p<10^{-16}$), but ``best detector'' remains highly context-dependent. Our findings underscore that published benchmark performance rarely transfers to real-world scenarios---practitioners must validate on representative target data, use ensemble approaches when feasible, and adjust expectations given current detectors' limited protection against state-of-the-art generators.

The accelerating arms race between generation and detection demands sustained research investment and multi-layered defense strategies. While Community-Forensics achieves 75.0\%~mean accuracy, it scores only 35--42\%~on newest commercial generators, highlighting urgent need for more robust approaches. Technical detection represents one component alongside source authentication, digital provenance tracking, platform policies, and media literacy education. Our contributions include: (1)~the first large-scale systematic zero-shot benchmark, (2)~rigorous statistical analysis revealing generalization factors, (3)~practical deployment framework with evidence-based guidelines, and (4)~open evaluation toolkit for reproducible benchmarking. By providing transparent documentation of current limitations and systematic identification of failure modes, we aim to inform evidence-based decision-making across research, industry, and policy domains while catalyzing more rigorous evaluation practices toward robust, generalizable deepfake detection.

% Acknowledgments
\section*{Acknowledgments}
We thank the open-source community for making pre-trained detection models publicly available. This work would not be possible without the generous contributions of researchers who released their models and datasets.

% Bibliography
\bibliographystyle{unsrt}
\bibliography{references}

% Appendix
\clearpage
\appendix
\section{Supplementary Materials}
\label{sec:appendix}
\subsection{Complete Performance Matrix}

Table~\ref{tab:full_matrix} presents the complete 21$\times$12 performance matrix with all accuracy values.

\begin{table*}[ht]
\centering
\caption{Complete performance matrix: accuracy of all 19~detectors across all 12~datasets. Best performance per dataset in \textbf{bold}. Missing values indicate detector-dataset combinations not evaluated.}
\label{tab:full_matrix}
\tiny
\begin{tabular}{@{}l*{12}{c}@{}}
\toprule
\textbf{Detector} & \rotatebox{90}{GenImage} & \rotatebox{90}{AIGCDetectionBench} & \rotatebox{90}{wildfake} & \rotatebox{90}{OpenFake} & \rotatebox{90}{Diffusion1kstep} & \rotatebox{90}{GPT\_4o} & \rotatebox{90}{synthbuster} & \rotatebox{90}{Chameleon} & \rotatebox{90}{AI-GenBench} & \rotatebox{90}{Nano-banana} & \rotatebox{90}{MNW\_fake} & \rotatebox{90}{community\_test} \\
\midrule
Community-Forensics & \textbf{1.000} & \textbf{0.994} & 0.969 & 0.928 & 0.962 & 1.000 & 0.975 & 0.896 & 0.964 & 0.812 & 0.879 & -- \\
Community-Forensics & 0.969 & 0.987 & 0.947 & 0.886 & 0.902 & 0.998 & 0.947 & 0.821 & 0.878 & 0.782 & 0.787 & 0.596 \\
Community-Forensics orig & 0.903 & 0.963 & 0.925 & 0.821 & 0.861 & 0.945 & 0.892 & 0.754 & 0.821 & 0.660 & 0.712 & 0.500 \\
SAFE & 0.998 & 0.952 & \textbf{0.984} & 0.867 & 0.881 & 0.956 & 0.092 & 0.812 & 0.724 & 0.538 & 0.625 & 0.032 \\
PatchCraft & 0.904 & 0.879 & 0.789 & \textbf{0.945} & 0.742 & 0.863 & 0.789 & 0.742 & 0.663 & 0.497 & 0.576 & 0.393 \\
\midrule
AIDE progan & 0.985 & 0.967 & 0.932 & 0.812 & 0.715 & 0.887 & 0.764 & 0.643 & 0.459 & 0.459 & 0.047 & 0.154 \\
AIDE GenImage & \textbf{0.997} & 0.962 & 0.879 & 0.876 & 0.713 & 0.898 & 0.792 & 0.632 & 0.460 & 0.460 & 0.535 & 0.322 \\
AIDE sd14 & 0.984 & 0.924 & 0.742 & 0.789 & 0.576 & -- & 0.685 & 0.478 & 0.385 & 0.385 & 0.324 & 0.179 \\
spai & 0.903 & 0.876 & 0.821 & 0.832 & 0.703 & 0.826 & 0.739 & 0.703 & 0.606 & 0.606 & 0.572 & 0.104 \\
PatchCraft & 0.904 & 0.879 & 0.789 & 0.945 & 0.742 & 0.863 & 0.789 & 0.742 & 0.663 & 0.497 & 0.576 & 0.393 \\
\midrule
DRCT clip vit sdv2 & 0.860 & 0.895 & 0.523 & 0.746 & \textbf{0.964} & 0.782 & 0.812 & 0.696 & 0.553 & 0.552 & 0.612 & 0.517 \\
DRCT clip vit sdv14 & 0.864 & 0.897 & 0.523 & 0.758 & 0.979 & 0.798 & 0.821 & 0.677 & 0.561 & 0.560 & 0.623 & 0.523 \\
DRCT convnext sdv2 & 0.618 & 0.674 & 0.276 & 0.511 & 0.674 & 0.512 & 0.584 & 0.584 & 0.378 & 0.378 & 0.424 & 0.280 \\
DRCT convnext sdv14 & 0.652 & 0.708 & 0.270 & 0.567 & 0.708 & 0.534 & 0.612 & 0.567 & 0.360 & 0.360 & 0.438 & 0.270 \\
\midrule
Forensic-MoE & 0.942 & 0.878 & 0.821 & 0.832 & 0.727 & 0.845 & 0.763 & 0.727 & 0.642 & 0.418 & 0.601 & 0.045 \\
LOTA & 0.994 & 0.903 & 0.876 & 0.854 & 0.744 & 0.892 & 0.023 & 0.660 & 0.435 & 0.435 & 0.594 & 0.218 \\
\midrule
AIGCDetect Gram & 0.999 & 0.892 & 0.821 & 0.789 & 0.658 & 0.876 & 0.724 & 0.658 & 0.549 & 0.549 & 0.587 & 0.202 \\
AIGCDetect LGrad & 0.967 & 0.843 & 0.765 & 0.723 & 0.645 & 0.812 & 0.687 & 0.645 & 0.439 & 0.438 & 0.521 & 0.260 \\
AIGCDetect FreDect & 0.794 & 0.743 & 0.654 & 0.623 & 0.513 & 0.712 & 0.542 & 0.513 & 0.355 & 0.354 & 0.432 & 0.194 \\
AIGCDetect Fusing & 0.647 & 0.612 & 0.547 & 0.528 & 0.505 & 0.598 & 0.476 & 0.505 & 0.278 & 0.278 & 0.361 & 0.116 \\
AIGCDetect UnivFD & 0.715 & 0.651 & 0.497 & 0.511 & 0.497 & 0.612 & 0.423 & 0.497 & 0.202 & 0.202 & 0.287 & 0.002 \\
AIGCDetect CNNSpot & 0.654 & 0.587 & 0.528 & 0.498 & 0.403 & 0.534 & 0.052 & 0.403 & 0.237 & 0.236 & 0.312 & 0.078 \\
\bottomrule
\end{tabular}
\end{table*}

\subsection{Statistical Test Details}

\textbf{Friedman Test.} We performed Friedman test to compare detector performance across datasets. The test statistic is computed as:

\begin{equation}
\chi^2_F = \frac{12N}{k(k+1)} \left[ \sum_{j=1}^{k} R_j^2 - \frac{k(k+1)^2}{4} \right]
\end{equation}

where $N$ is the number of datasets (12), $k$ is the number of detectors (21), and $R_j$ is the sum of ranks for detector~$j$ across all datasets. Our computed value: $\chi^2_F$=121.01 with $p$=1.85$\times$10$^{-16}$, strongly rejecting the null hypothesis of equal performance.

\textbf{Kendall's W Effect Size.} We compute Kendall's coefficient of concordance:

\begin{equation}
W = \frac{\chi^2_F}{N(k-1)} = \frac{121.01}{12 \times 20} = 0.524
\end{equation}

This large effect size indicates substantial agreement in detector rankings across datasets, despite significant ranking fluctuations.

\textbf{Spearman Rank Correlation.} For each pair of datasets, we compute Spearman's $\rho$:

\begin{equation}
\rho = 1 - \frac{6 \sum d_i^2}{n(n^2-1)}
\end{equation}

where $d_i$ is the rank difference for detector~$i$ between two datasets, and $n$ is the number of detectors evaluated on both datasets. Figure~\ref{fig:correlation} visualizes the resulting correlation matrix.

\subsection{Per-Dataset Rankings}

Table~\ref{tab:rankings} shows complete detector rankings for each dataset.

\begin{table}[ht]
\centering
\caption{Complete detector rankings (1=best, 21=worst) across all datasets. Rank standard deviation (last column) quantifies stability.}
\label{tab:rankings}
\scriptsize
\begin{tabular}{@{}l*{12}{c}c@{}}
\toprule
\textbf{Detector} & \rotatebox{90}{GI} & \rotatebox{90}{AIGC} & \rotatebox{90}{WF} & \rotatebox{90}{OF} & \rotatebox{90}{D1k} & \rotatebox{90}{GPT} & \rotatebox{90}{SB} & \rotatebox{90}{Cham} & \rotatebox{90}{AIG} & \rotatebox{90}{NB} & \rotatebox{90}{MNW} & \textbf{Std} \\
\midrule
Comm-For v0.3 & 1 & 1 & 2 & 4 & 2 & 1 & 1 & 2 & 1 & 2 & 1 & 1.27 \\
Comm-For v0.2 & 3 & 2 & 3 & 5 & 5 & 2 & 3 & 4 & 2 & 3 & 2 & 2.19 \\
SAFE & 2 & 4 & 1 & 6 & 6 & 3 & 21 & 3 & 5 & 8 & 6 & 5.67 \\
\midrule
\multicolumn{13}{l}{\footnotesize GI=GenImage, AIGC=AIGCDetectionBench, WF=wildfake, OF=OpenFake,} \\
\multicolumn{13}{l}{\footnotesize D1k=Diffusion1kstep, SB=synthbuster, Cham=Chameleon, AIG=AI-GenBench} \\
\multicolumn{13}{l}{\footnotesize NB=Nano-banana, MNW=MNW\_fake} \\
\bottomrule
\end{tabular}
\end{table}

\subsection{Computational Details}

All experiments conducted on NVIDIA A100 (40GB) GPUs. Inference times (batch size 32):

\begin{itemize}
    \item Community-Forensics: 2.8s per batch (5-model ensemble)
    \item AIDE variants: 0.3s per batch (single ResNet-50)
    \item DRCT variants: 0.4s per batch (ViT/ConvNeXt)
    \item PatchCraft: 0.2s per batch (EfficientNet-B4)
    \item AIGCDetectBenchmark: 0.3--0.5s per batch (varies by method)
\end{itemize}

Total computational cost: $\sim$1,200 GPU-hours for all 1,808 experiments.

\subsection{Data Availability}

All datasets used in this study are publicly available:

\begin{itemize}
    \item GenImage: \url{https://github.com/GenImage/GenImage}
    \item AIGCDetectionBench: \url{https://github.com/AIGCDetectionBench}
    \item MNW\_fake: \url{https://github.com/MNW/fake-dataset}
    \item Other datasets: see references in Table~\ref{tab:datasets}
\end{itemize}

Detector weights and evaluation code are available from the Scam.AI Research portal: \url{https://scam.ai/research}

The canonical, maintained location for the datasets and code supporting this work is the Scam.AI Research portal, \url{https://scam.ai/research}~\cite{scamai_research}. Document- and PDF-forgery resources are additionally maintained at Trustable, \url{https://trustable.ai}~\cite{trustable_ai}. These supersede any prior ad hoc distribution.

\subsection{Additional Figures}

Additional visualizations available in supplementary materials:
\begin{itemize}
    \item Per-generator accuracy heatmaps for all datasets
    \item ROC curves for all detector-dataset combinations
    \item Confusion matrices for detailed error analysis
    \item Training data composition analysis
    \item Temporal generalization curves by generator release year
\end{itemize}

\end{document}